\documentclass[graybox]{svmult}

\usepackage{type1cm}        
\usepackage{makeidx}         
\usepackage{graphicx}        
\usepackage{multicol}        
\usepackage[bottom]{footmisc}

\usepackage{newtxtext}  %
\usepackage{newtxmath}  

\begin{document}

\author{%
Beatrice Portelli$^*$%
, Daniele Passab\`i$^*$%
, Edoardo Lenzi$^*$%
, Giuseppe Serra$^*$%
,\\ Enrico Santus$^\dagger$%
\ and Emmanuele Chersoni$^\ddagger$%
\\[.2cm]
\textnormal{\footnotesize
$^*$ University of Udine, Udine (UD), Italy\\
$^\dagger$ CSAIL MIT, Cambridge (MA), United States of America\\
$^\ddagger$ The Hong Kong Polytechnic University, Hong Kong\\[.2cm]
portelli.beatrice@spes.uniud.it\\
passabi.daniele@spes.uniud.it\\
lenzi.edoardo@spes.uniud.it\\
giuseppe.serra@uniud.it\\
esantus@csail.mit.edu\\
emmanuele.chersoni@polyu.edu.hk\\
}
}

\title*{Improving Adverse Drug Event Extraction with SpanBERT on Different Text Typologies}
\label{chap:improving_ade_extraction}
\titlerunning{Improving ADE Extraction with SpanBERT on Different Text Typologies}

\maketitle

\abstract*{In recent years, Internet users are reporting Adverse Drug Events (ADE) on social media, blogs and health forums. Because of the large volume of reports, pharmacovigilance is seeking to resort to NLP to monitor these outlets.
We propose for the first time the use of the SpanBERT architecture for the task of ADE extraction: this new version of the popular BERT transformer showed improved capabilities with multi-token text spans. We validate our hypothesis with experiments on two datasets (SMM4H and CADEC) with different text typologies (tweets and blog posts),
finding that SpanBERT combined with a CRF outperforms all the competitors on both of them.}

\abstract{In recent years, Internet users are reporting Adverse Drug Events (ADE) on social media, blogs and health forums. Because of the large volume of reports, pharmacovigilance is seeking to resort to NLP to monitor these outlets.
We propose for the first time the use of the SpanBERT architecture for the task of ADE extraction: this new version of the popular BERT transformer showed improved capabilities with multi-token text spans. We validate our hypothesis with experiments on two datasets (SMM4H and CADEC) with different text typologies (tweets and blog posts),
finding that SpanBERT combined with a CRF outperforms all the competitors on both of them.}

\section{Introduction}

Regulators, such as the Food and Drug Administration (FDA) and the European Medicine Agency (EMA), approve every year dozens of drugs, after verifying their safety in clinical trials. Sometimes, however, clinical trials are not sufficient to discover all potential Adverse Drug Events (ADE). Pharmacovigilance, therefore, monitors the drugs in the market to ensure that unexpected effects are timely identified and actions are taken to minimize their harm.
A constantly growing number of patients chooses to describe the side effects on social media platforms, health forums and similar outlets.
The growing amount of available ADE information in such sources and the high cost of manual extraction have recently made the automation of ADE extraction an important research topic in the natural language processing community (NLP), also encouraging a dedicated series of shared tasks \cite{sarker2017overview,weissenbacher2018overview,weissenbacher2019overview}. Such extraction poses two major challenges, namely (i) the identification of relevant content among a large amount of unrelated posts; and (ii) the understanding of symptoms even when they are expressed in layman's terms.

In this paper we address the automation of ADE extraction through a tagging task: given a text, our goal is to identify whether it contains an ADE and its precise span. Because this task is often propaedeutic to other tasks (e.g. the mapping of the extracted ADEs to medical ontologies: see \cite{weissenbacher2019overview}) it is of key importance that the relevant span of tokens is identified as precisely as possible.

To this end, we apply for the first time SpanBERT \cite{Joshi2019SpanBERTIP}, a recently proposed version of BERT \cite{Devlin2019BERTPO} which has shown improved capabilities on multi-token text spans, such as those describing ADE. The performance of the system is evaluated on two widely-used benchmark datasets, containing two different kinds of texts: (i) short and highly informal tweets, in SMM4H \cite{weissenbacher2019overview}; and longer and more structured blog posts in CADEC \cite{Karimi2015CadecAC}. In our experiments, we prove that this approach obtains competitive performance not only compared to common baselines but also compared to state-of-the-art transformer-based architectures. Moreover, our results show that, when SpanBERT is combined with CRF, it outperforms all the other approaches on both types of texts.

\section{Related Work}

The extraction of ADE mentions from social media texts is a task which started receiving growing attention in the last few years. This was prompted by the increasing number of users who discuss their drug-related experiences online, in platforms such as Twitter. Studies like \cite{Sarker2015PortableAT,Nikfarjam2015PharmacovigilanceFS,daniulaityte2016bad} were among the first to propose machine learning approaches for this task, either using traditional feature engineering or word embeddings-based solutions.

The problem gained visibility with the introduction of the Social Media Mining for Health Applications (SMM4H) Workshop and the shared tasks co-located with it.
Methods based on neural networks became a more common choice \cite{wu2018detecting,nikhil2018neural}.
During the latest instance of SMM4H, co-located with ACL 2019, transformers-based architectures such as BERT \cite{Devlin2019BERTPO} and BioBERT \cite{lee2020biobert} were the building blocks of the top performing systems \cite{chen2019hitsz,mahata2019midas,miftahutdinov2019kfu}.
However, these recent shared tasks have focused on a very particular type of social media texts: tweets, which are short, highly informal and noisy texts. This led the research community to pay less attention to longer and more structured social media texts, like forum posts, which are another interesting text typology for this kind of task.

The most recent progresses on non-tweet based benchmarks include thoughtful analyses of ADE extraction framed as a Named Entity Recognition task with discontinuous entity mentions (see \cite{dai2018recognizing} for a complete overview). Such studies focus on the fact that the ADE mentions might be long and include unconventional descriptions, and they typically evaluate their approaches on more structured text typologies (i.e. texts from medical forums) \cite{stanovsky2017recognizing,dai2020EffectiveTransition}, while the issue of their performance on short and informal texts has not been addressed, to the best of our knowledge.

\section{The Proposed Approach}
\label{sec:approach}

\begin{figure}[t]
\centering
\includegraphics[width=\linewidth]{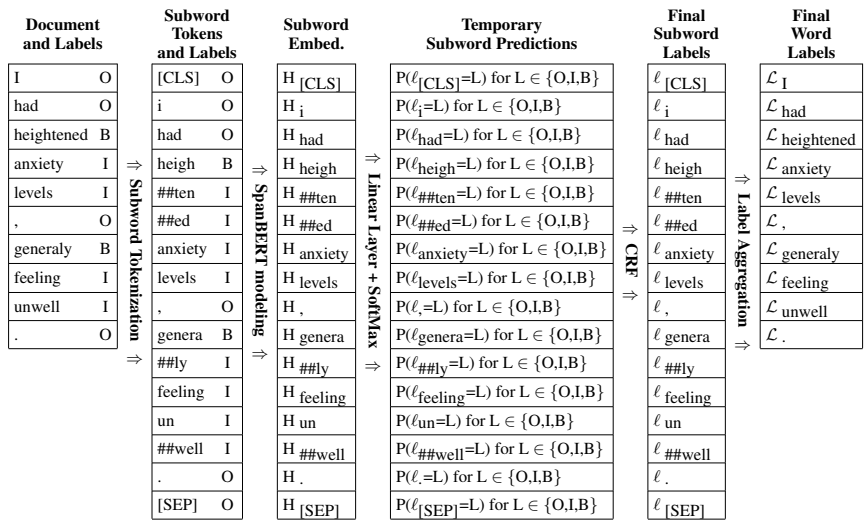}
\caption{Schema of the proposed SpanBERT+CRF architecture.}
\label{fig:architecture}
\end{figure}

Pre-trained methods, such as BERT \cite{Devlin2019BERTPO} and BioBERT \cite{lee2020biobert} have recently shown high gains in the ADE extraction task compared to previous approaches. However, these algorithms are trained in a self-supervised fashion by masking individual words or subwords units. In the ADE extraction task, longer spans and relations between multiple spans of text are instead the key.
In addition, in short and highly contextual language, such as the one used in social media -- which is characterized by acronyms, slang, metaphors, etc. -- the meaning of each word in strongly influenced by the ones in its vicinity. 

SpanBERT \cite{Joshi2019SpanBERTIP} takes into account wider portions of text by adopting both a different masking scheme and a different training objective. In fact, during training, random contiguous spans of tokens are masked, rather than individual words, forcing the model to predict the full span from the known tokens at its boundaries (the span-boundary objective - SBO). This basically means that the hidden representation of each word is encoding span-level information for its neighbours.

Since the SpanBERT model, to the best of our knowledge, has never been tested on the ADE extraction task \footnote{Notice however that an architecture inspired by similar principles, SpERT, has been recently tested on ADE entity extraction and on the classification of drug-reaction relation, leading to improvements on both tasks  \cite{eberts2020span}.}, we propose a neural network architecture that combines SpanBERT with a Conditional Random Field classifier (henceforth CRF) \cite{lafferty2001ConditionalRF,papay2020dissecting}.

Each sample goes through five main processing steps, which lead to the extraction of the ADE entities contained within the text. The following paragraphs describe these steps in detail, while Fig. \ref{fig:architecture} shows a schema of the architecture.

\runinhead{Preprocessing}

Each training sample is originally accompanied by a list of tuples which record the start- and the end-character of each ADE mention. As a first preprocessing step, this information is converted into the commonly used IOB (Inside-Begin-Outside) annotation schema. More precisely, each sentence is split into words and each word is labelled with one of the three letters: \texttt{B} if it is the beginning (first word) of an ADE mention; \texttt{I} if it is the continuation of an ADE mention; \texttt{O} if the word is outside of any ADE mention.
For example, if we consider the sentence ``I had heightened anxiety levels, generaly feeling unwell.'' (analyzed in Fig. \ref{fig:architecture}) the phrase ``heightened anxiety levels'' is an ADE mention made of three words, which gets labeled as [\texttt{B}, \texttt{I}, \texttt{I}] respectively.

\runinhead{Subword Tokenization}
BERT-based models usually employ a tokenization method called Wordpiece tokenization \cite{Wu2016GooglesNM}, which allows to tokenize any word without incurring into ``out-of-vocabulary'' tokens. In particular, the vocabulary of the tokenizer is composed both of full words and wordpieces (sub-word units, which can be as short as single characters). Even if a word is not directly part of the vocabulary, it can always be split into a set of sub-words which belong to the vocabulary. For example the word ``heightened'' is not part of the SpanBERT vocabulary, but it can be split into three subwords: [``heigh'', ``\#\#ten'', ``\#\#ed''] (wordpieces can be identified by the leading \#\#). Since each full word was labelled using the IOB schema, we need to decide how to assign the labels to the eventual subword tokens.
We choose to generate sub-labels which are consistent with the IOB schema: words labelled as \texttt{B} generate a series of subwords labelled as [\texttt{B}, \texttt{I}, \dots, \texttt{I}], while words labelled as \texttt{I} (or \texttt{O}) generate a series of identical \texttt{I} (or \texttt{O}) sub-labels.

In addition to the Wordpiece tokenization, BERT-base models usually employ additional special tokens to prepare the sentence before feeding it to the model: a \texttt{[CLS]} (classification) token that marks the beginning of the sample; a \texttt{[SEP]} (separator) token to mark the end of the sample; a \texttt{[PAD]} (padding) token to pad all input sequences to the same length (not shown in the image). All predictions generated on the \texttt{[PAD]} tokens are discarded and are not taken into consideration when computing the loss for the model.

\runinhead{SpanBERT modeling}

The first actual component of the architecture if a SpanBERT neural network, composed of 12 encoder layers with multi-head attention. The network takes as input the sequence of subword tokens $[x_i]$ and outputs a sequence of embeddings $[\textrm{H}_{x_i}]$ (one for each subword token). Internally, each one of the transformer (encoder) layers computes an intermediate embedding of each token. The attention mechanism allows the network to enrich the embeddings with contextual information after each layer.

\runinhead{Intermediate label prediction}

The embedding of each token is processed by a linear layer, which projects the embedding of each subword token to a vector of size 3 (representing the IOB labels for the tokens). The softmax operator turns the raw value into a probability distribution over the possible output labels.

\runinhead{Conditional Random Field}

The CRF module takes as input the probability distribution generated by the SpanBERT model, and produces another sequence of subword-level IOB labels. This step aims at denoising the output labels produced by the previous components exploiting a well known statistical model, which was widely used for this task prior to the introduction of deep neural networks.

\runinhead{Label aggregation}

Finally, the subword-level predictions need to be aggregated to produce word-level labels which can be easily mapped to the original input sequence.

A set of subword labels $\{\ell_i\}$, is aggregated into a word label $\mathcal{L}$ using the first rule that applies:
(i) if $\ell_i=\texttt{O}$ for all $i$, then $\mathcal{L}=\texttt{O}$;
(ii) if $\ell_i=\texttt{B}$ for any $i$, then $\mathcal{L}=\texttt{B}$;
(iii) if $\ell_i=\texttt{I}$ for any $i$, then $\mathcal{L}=\texttt{I}$.
The aggregated output is therefore a sequence of word-level IOB labels, which can be compared to preprocessed inputs or easily converted into character level indices.

\section{Experimental Results}

\subsection{Datasets}

The two datasets chosen for the experiments two widely used ADE extraction datasets.
{SMM4H} is the training dataset for Task 2 of the Shared Task SMM4H 2019 \cite{weissenbacher2019overview}. It contains 2276 tweets which mention at least one drug name (1300 tweets are positive for the presence of ADEs while the other 976 are negative control samples). We further partitioned the training set into training and validation sets (80:20), maintaining the proportions of positive and negative samples.
The competition also includes a blind test set composed of 1573 unlabelled tweets.
The final metrics are evaluated on the blind test set using the scoring system provided by the competition\footnote{\url{https://competitions.codalab.org/competitions/20798}}. This ensures that the results are comparable with all the systems which took part in the shared task.
{CADEC} \cite{Karimi2015CadecAC}
is a widely used and publicly available ADE extraction dataset. It contains 1250 posts from the health-related forum ``AskaPatient'', annotated for the presence of ADEs. We use the same training, validation and test splits made publicly available by \cite{dai2020EffectiveTransition}.

\subsection{Metrics}
As evaluation metrics we use the Strict and Partial (also known as Relaxed) F1-Score, which are commonly adopted for this task \cite{Semeval2013}. They are computed on entity level (as opposed to the token level).
The partial score takes into account ``partial'' matches, in which it is sufficient for a prediction to partially overlap with the gold annotation. On the other hand, the Strict score requires a perfect match between the system output and the annotated ADE.

\subsection{Baselines}

We directly compare our proposed SpanBERT+CRF on the two datasets with four different architectures.

{BERT} (version base, uncased)\footnote{Preliminary experiments with cased and uncased versions showed that the latter performed better.}, a Transformers-based baseline  finetuned on the training data. The outputs of the final hidden layer are passed to a linear layer which projects the hidden representation to the needed output space (one label per token). 

{BERT+CRF} the same BERT model but with a CRF classifier on top, used to denoise the output of BERT as we described in Section \ref{sec:approach}.

{SpanBERT} (version base, cased) is the simpler version of the proposed model, with only a linear layer on top.

{TMRLeiden} \cite{Dirkson2019TransferLF}: 4th classified in the ADE-extraction task at the SMM4H 2019 Workshop \cite{weissenbacher2019overview}. It is composed of a BiLSTM network taking as input a concatenation of BERT embeddings and Flair embeddings \cite{akbik2019flair}.
It was chosen for direct comparison since the first four systems in the competition achieved comparable results (TMRLeiden had the second best Strict F1-Score) and TMRLeiden was the only top-team which made their code public, which allows us to compare their system with others (and on different datasets) with only minimal adjustments.

We perform a partial comparison with the top-3 classified at SMM4H 2019 (\textit{KFU NLP} \cite{miftahutdinov2019kfu}, {THU\_NGN} \cite{ge-etal-2019-detecting}, {MIDAS@IHTD} \cite{mahata2019midas}) by reporting their scores on the SMM4H dataset as published in the overview of the workshop \cite{weissenbacher2019overview}.

\subsection{Implementation Details}

TMRLeiden was reimplemented using its original code as a base\footnote{\url{https://github.com/AnneDirkson/SharedTaskSMM4H2019}}
and trained according to the details in its paper.

For the transformer-based architectures
we performed a parameter-tuning analysis via grid-search using different learning rates
($[5e{-4}, 5e{-5}, 5e{-6}]$)
and dropout rates (from $0.15$ to $0.30$, increments of $0.05$).
All the architectures were trained for $50$ epochs on the training set. The best learning rate, dropout rate and maximum epoch were chosen evaluating the models on the validation set and using the Partial F1-score as the performance metric.
The models were trained using the best hyperparameters on the concatenation of the training set and the validation set. This procedure was repeated five times with different random seeds and the results on the test set have been averaged.

\subsection{Evaluation}

\begin{table}[!t]
\caption{\label{tab:results} Strict and Partial F1-Score with standard deviations for all architectures on both datasets}
\centering
\begin{tabular}{p{3cm}*{4}{p{2cm}}}
\hline\noalign{\smallskip}
Architecture &
Partial F1 \newline CADEC &
Partial F1 \newline SMM4H &
Strict F1 \newline CADEC &
Strict F1 \newline SMM4H \\
\noalign{\smallskip}\svhline\noalign{\smallskip}
BERT &
{77.7 $\pm$ 0.3} &
{68.7 $\pm$ 0.4} &
{65.2 $\pm$ 0.5} &
{43.5 $\pm$ 0.4} \\
\textbf{BERT+CRF} &
{77.2 $\pm$ 0.5} &
\textbf{69.8 $\pm$ 0.2} &
{64.4 $\pm$ 0.8} &
{45.8 $\pm$ 0.9} \\
SpanBERT &
{79.2 $\pm$ 0.6} &
{65.8 $\pm$ 2.2} &
{67.2 $\pm$ 0.8} &
{47.1 $\pm$ 1.9} \\
\textbf{SpanBERT+CRF} &
\textbf{79.4 $\pm$ 0.3} &
{67.7 $\pm$ 0.7} &
\textbf{67.6 $\pm$ 0.6} &
\textbf{47.5 $\pm$ 0.2} \\
\noalign{\smallskip}\hline\noalign{\smallskip}
TMRLeiden {\cite{Dirkson2019TransferLF}}&
{77.1 $\pm$ 0.8} & 62.5 &
{65.0 $\pm$ 1.1} & 43.1 \\ 
KFU NLP {\cite{miftahutdinov2019kfu}}&
\multicolumn{1}{c}{--}  & 65.8 &
\multicolumn{1}{c}{--}  & 46.4 \\ 
THU\_NGN {\cite{ge-etal-2019-detecting}}&
\multicolumn{1}{c}{--}  & 65.3 &
\multicolumn{1}{c}{--}  & 35.6 \\ 
MIDAS@IIITD {\cite{mahata2019midas}}&
\multicolumn{1}{c}{--}  & 64.1 &
\multicolumn{1}{c}{--}  & 32.8 \\ 
\noalign{\smallskip}\hline\noalign{\smallskip}
\end{tabular}
\end{table}

Table \ref{tab:results} presents a comparative evaluation between our architecture and the other approaches on both datasets (we reported the scores of the shared task 2019 for the systems whose code is not publicly available).
Results are partitioned into two sections, one for each metric.

Looking at the results on CADEC, we can see that SpanBERT+CRF achieves the best performance for both metrics, showing that our approach is able to properly deal with the long texts contained in forum posts.
SpanBERT (without CRF) gets the second place, while TMRLeiden achieves a performance slightly lower than BERT, and with an higher standard deviation. This could be explained by the fact that TMRLeiden was specifically designed to perform this task on tweets, and does not express its full potential on forum posts. This finding confirms that it is worth testing ADE systems on different textual typologies, since text genre has an important impact on the performance.

Considering the results on the SMM4H dataset: we see that the proposed SpanBERT+CRF achieves the best performance on the Strict metric, while BERT-based architectures perform better on the Partial metric. The BERT-based variants only surpass SpanBERT+CRF in this particular case, and only on the tweets data. Interestingly, all these simple architectures improve over the performances of all the top-4 systems that took part in the SMM4H 2019 shared task.

More in general, models tend to perform better on CADEC. This is most likely due to the larger size of the dataset and to the more regular type of text, compared to the noisy tweets.
We can also notice that the denoising action of the CRF seems to be effective in most situations (e.g. the scores for the simple SpanBERT are lower than the ones achieved by SpanBERT+CRF).

We take this as a preliminary evidence that SpanBERT is a particularly convenient solution for dealing with ADE, especially for longer and more structured texts, and that it is more precise in detecting the ADE boundaries, as suggested by the more marked improvements for the Strict metric.

\subsection{Quantitative Analysis}

We performed a series on quantitative tests comparing the sets of ADE entities predicted by the various models. The objective was to investigate whether the different architectures produce significantly different results.

\runinhead{True Positive and False Positive Distribution}

\begin{table}[!t]
\caption{\label{tab:mcnemar}
p-values for the McNemar test on all pairs of architectures on the two datasets}
\centering
\begin{tabular}{*{2}{p{2.5cm}}*{2}{p{1.5cm}}}
\hline\noalign{\smallskip}
Architecture 1 & Architecture 2 & SMM4H & CADEC \\
\noalign{\smallskip}\svhline\noalign{\smallskip}
BERT & BERT+CRF          & \textbf{< 0.001} & \textbf{< 0.001} \\
BERT & SpanBERT          & \textbf{< 0.001} & $\geq$ 0.1       \\
BERT & SpanBERT+CRF      & \textbf{< 0.001} & $\geq$ 0.1       \\
BERT & TMRLeiden         & \textbf{< 0.001} & \textbf{< 0.001} \\
BERT+CRF & SpanBERT      & \textbf{< 0.001} & \textbf{< 0.001} \\
BERT+CRF & SpanBERT+CRF  & $\geq$ 0.1       & $\geq$ 0.1       \\
BERT+CRF & TMRLeiden     & \textbf{< 0.001} & \textbf{< 0.001} \\
SpanBERT & SpanBERT+CRF  & \textbf{< 0.001} & $\geq$ 0.1       \\
SpanBERT & TMRLeiden     & \textbf{< 0.01}  & \textbf{< 0.001} \\
SpanBERT+CRF & TMRLeiden & \textbf{< 0.001} & \textbf{< 0.001} \\
\noalign{\smallskip}\hline\noalign{\smallskip}
\end{tabular}
\end{table}

First of all we compared the true positive and false positive predictions for all the models using the McNemar test. Since the gold labels for the SMM4H test set are not public, we compared the predictions generated on the validation set (training all the models on the training set only).

The p-values obtained by comparing each pair of models are summarized in Table \ref{tab:mcnemar}.
The differences are statistically significant (in most of the cases with $p < 0.001$) on the SMM4H dataset, with the exception of the comparison between SpanBERT+CRF and BERT+CRF. Indeed, even if the two models are different in terms of overall strict F1 score (as shown in Table \ref{tab:results}), they are really close in terms of precision and mainly differ due to BERT+CRF having a lower recall.
The situation is less defined on the CADEC dataset, with the differences being clearly significant only when comparing BERT-based-models with the baseline TMRLeiden. This is due to all these models reporting similar precision values on CADEC (while they have different recall values, which determined the differences in the final F1 scores in Table \ref{tab:results}).

\runinhead{Text Statistics of the Predictions}

\begin{table}[!t]
\caption{\label{tab:text_metrics}
Average text metrics for the BERT-based models on the SMM4H dataset}
\centering
\begin{tabular}{p{3.2cm}*{3}{p{1.9cm}}p{2.2cm}}
\hline\noalign{\smallskip}
& BERT & BERT+CRF & SpanBERT & SpanBERT+CRF \\
\noalign{\smallskip}\svhline\noalign{\smallskip}
{Dale Chall Readability$^+$} &
\textbf{{08.34 $\pm$ 0.24}} & {08.15 $\pm$ 0.25} &
{08.20 $\pm$ 0.22} & {08.32 $\pm$ 0.24} \\
{Automated Readability$^+$}  &
{08.42 $\pm$ 0.31} & {08.37 $\pm$ 0.31} &
{08.35 $\pm$ 0.28} & \textbf{{08.51 $\pm$ 0.30}} \\
{Flesch Reading Ease$^-$}    &
{62.73 $\pm$ 1.98} & {63.57 $\pm$ 2.08} &
\textbf{{62.60 $\pm$ 1.82}} & {62.92 $\pm$ 1.97} \\
{Syllable Count$^+$}        &
{02.49 $\pm$ 0.04} & {02.46 $\pm$ 0.04} &
\textbf{{02.89 $\pm$ 0.05}} & {02.79 $\pm$ 0.05} \\
{Character Length$^+$}              &
{09.66 $\pm$ 0.14} & {09.60 $\pm$ 0.15} &
\textbf{{11.29 $\pm$ 0.18}} & {10.94 $\pm$ 0.18} \\
\noalign{\smallskip}\hline\noalign{\smallskip}
\end{tabular}\\
$^+$ Higher means less common  \hspace{.5cm}
$^-$ Lower means less common
\end{table}

We also computed a series of text-specific metrics on all the predictions generated by the models on the test data.

The chosen metrics are:
three widely used readability indices (Dale–Chall readability score, Automated readability index, Flesch–Kincaid reading ease)
and two measures of text length (Syllable Count and Character Length).
The readability indices indicate whether the extracted text contains words which are common or easy to read in the English language. In this case the presence of uncommon or difficult-to-read words could be indicative of the use of medical terms or uncommon expressions in the predicted entities.
Measuring the length of the whole predictions by syllables or by characters is also an indicator of the complexity of the terms and the ability of the model to predict significant contiguous spans of text.

Table \ref{tab:text_metrics} summarizes the metrics for all the architectures. We can see that the predictions generated by the SpanBERT-based models show an higher frequency of uncommon (or difficult-to-read) words for all the metrics: lower readability scores and longer predictions.
To test whether these differences were significant
we used the non parametric Mann-Whitney-U significance test on each pair of models.
The test shows that:

\begin{enumerate}
\item{when comparing two architectures based on the same model (e.g. BERT with and without CRF) differences are not significant, for all the considered metrics. This means that the addition of the CRF enhances the performances of the model without altering the textual style of the predictions.}

\item{when comparing BERT-based architectures with SpanBERT-based architectures Syllable Count and Character Length are significantly different ($p < 0.001$). This means that using a model which was pretrained with span-level knowledge affects the length of the final predictions.}

\item{when comparing BERT-based architectures with SpanBERT-based architectures Dale Chall Readability and Flesch Reading Ease are significantly different ($p < 0.05$). In particular SpanBERT+CRF predicts entities containing less common words compared with the predictions of BERT+CRF.}
\end{enumerate}

Both analyses lead to the conclusion that the predictions of the architectures are significantly different, and that SpanBERT-based models predict longer entity mentions containing uncommon or ``difficult'' words.

\subsection{Qualitative Error Analysis}

\begin{table}[!t]
\caption{\label{tab:examples}
Examples of errors: gold annotations bold and enclosed in squared brackets, entities extracted by SpanBERT+CRF overlined, by TMRLeiden underlined}
\centering
\begin{tabular}{ll}
\hline\noalign{\smallskip}
\# & Text \\
\noalign{\smallskip}\svhline\noalign{\smallskip}
1 &
$[$\textbf{\underline{$\overline{\mbox{short term memory affected}}$}}$]$, $[$\textbf{$\overline{\mbox{\underline{loss of concentration} \underline{levels}}}$}$]$, $[$\textbf{$\overline{\mbox{\underline{dizzy}}}$}$]$,\\
\noalign{\smallskip}
& $[$\textbf{$\overline{\mbox{\underline{hightened} \underline{anxiety levels}}}$}$]$, $[$\textbf{$\overline{\mbox{\underline{generaly feeling unwell like walking in a daze} \underline{(zombie like)}}}$}$]$\\
\noalign{\smallskip}
& $[$\textbf{$\overline{\mbox{\underline{weakness in leg muscles}}}$}$]$.\\
\noalign{\smallskip}\hline\noalign{\smallskip}
2 &
this lozenge is making my $\overline{\mbox{whole mouth $[$\textbf{\underline{numb}}$]$}}$... is that normal? \\
\noalign{\smallskip}\hline\noalign{\smallskip}
3 &
user fluoxetine and quet combo $[$\textbf{$\overline{\mbox{zombified}}$}$]$ me... ah, the meds merrygoround bipolar \\
\noalign{\smallskip}\hline\noalign{\smallskip}
4 &
it feels \underline{like my} $[$\textbf{$\overline{\mbox{\underline{brain is melting}}}$}$]$. numbermg of vyvanse in number hours... \\
\noalign{\smallskip}
& not smart danielle. not smart. \\
\noalign{\smallskip}\hline\noalign{\smallskip}
5 &
i have a \underline{$\overline{\mbox{slipped vertabrae}}$} and a \underline{degenerative disk}. this medicine has been extrememly helpful. \\
\noalign{\smallskip}\hline\noalign{\smallskip}
6 &
i never had \underline{$\overline{\mbox{bleeding}}$} or \underline{$\overline{\mbox{vomiting}}$} just alot alot of $[$\textbf{\underline{$\overline{\mbox{fatigue}}$}}$]$ and face was $[$\textbf{\underline{$\overline{\mbox{pale}}$}}$]$ and \\
\noalign{\smallskip}
& $[$\textbf{\underline{$\overline{\mbox{lost tons of weight}}$}}$]$. \\
\noalign{\smallskip}\hline\noalign{\smallskip}
\end{tabular}
\end{table}

To illustrate the differences between our SpanBERT+CRF solution and a tweets-focused architecture such as TMRLeiden, we propose a qualitative error analysis (see Table \ref{tab:examples}). 
We selected the samples on which one of the architectures performed clearly better than the other one in terms of Strict F1-Score, and analyzed the results manually.
In particular, we observe that our solution has the tendency to extract longer ADE spans compared to TMRLeiden, and it is better at dividing consecutive ADE spans (Table \ref{tab:examples}, Example 1). This leads to better results on the CADEC datasets, which contains longer descriptions, but a worse precision on SMM4H, where the annotations are often one-word long (Example 2).
However, sometimes this is more a matter of annotation choice (``whole mouth numb'' in Example 2 is actually a more precise description of the ADE) and it impacts only the Strict evaluation metric.
SpanBERT+CRF seems also able to identify highly informal/slang descriptions (``zombified'', ``jacked up'', ``brain melting'', etc.), whereas TMRLeiden completely ignores them (Examples 3,4).

As regards false positives, both models make similar errors, misclassifying as ADE some mentions of diseases (Example 5), symptoms generated by the diseases themselves and repercussions of the ADE (e.g. extracting both ``joint pain in shoulder'', the ADE, and ``could not lift my arms over my head'', which is just a consequence of it).
This problem might be addressed by adding more output classes to the models, training them to jointly detect ADEs, Diseases, and Symptoms that are not connected to Drugs.

For both models, further errors are generated by the lack of proper handling of negation, and therefore samples reporting the absence of a specific ADE are erroneously treated as mentions of such ADE (Example 6).

\section{Conclusions}
We introduced the SpanBERT architecture to the task of ADE extraction from different types of texts. 
We carried out our experiments on two widely-used datasets, respectively containing informal tweets and longer blog posts, SpanBERT obtained competitive results for both Strict and Partial F1-score with state-of-the-art models, outperforming all of them in combination with a CRF for the Strict metric.
The error analysis also suggests that SpanBERT is more capable of modeling the presence of uncommon words and longer entity mentions when compared to the other approaches.
Our results show the flexibility of the model on different typologies of data, and its higher accuracy in modeling the span boundaries of the ADEs.


\bibliographystyle{spmpsci}
\bibliography{main}

\end{document}